# INDOOR NAVIGATION USING INFORMATION FROM A MAP AND A RANGEFINDER

O.A. STEPANOV[1],
Concern CSRI Elektropribor, ITMO University, Saint Petersburg, Russia.

MOSTAFA MANSOUR[2]
ITMO University, Saint Petersburg, Russia. mostafa.e.mansour@gmail.com

***Abstract***

*The problem of indoor navigation of mobile objects, using a map and measurements of distances to the walls is considered. A nonlinear filtering problem aimed at calculating the optimal, in the root-mean-square sense, of the sought parameters is formulated in the context of the Bayesian approach. The algorithm for its solution based on the point-mass method is described. The simulation results illustrating the advantages of the proposed problem statement and the resultant algorithm are discussed.*

## Introduction

In recent years, considerable attention has been given to indoor navigation of mobile objects in the conditions where the GNSS signals are unavailable [1-7]. In this case, the source of the information may be the measurements from various sensors, such as cameras, rangefinder, etc., and a plan of a building or a room with information on certain position of the walls, doors, window frames, and obstacles in a specified coordinate system. This problem is of nonlinear nature, which is due to nonlinear dependence of the measured parameters on the object's position and heading. In order to correctly account for this dependence, it is necessary to use nonlinear filtering methods [2-5]. However, the researchers are mainly aimed at obtaining efficient, from the computational standpoint, algorithms for determining the state vector being estimated. At the same time, the problem solution, in the context of nonlinear filtering theory, based on the Bayesian approach, offers various additional possibilities, which, in the authors' opinion, have not been fully taken into account. In particular, the algorithm may be designed so that, in addition to the estimate, it is also possible to form a current accuracy characteristic as a conditional covariance matrix of estimation errors. This is very important in integrated processing of navigation measurements. Given such characteristics, it becomes possible to correctly account for the accuracy of estimates subjected to integrated processing and, as a result, increase the efficiency of this processing. In addition, in this case we can estimate the efficiency of using various measurements to solve the problem and thus to plan which of the measurements should be used, in order to reduce the computational load. It is the analysis of these additional possibilities in solving the indoor navigation problem that this paper is devoted to. We consider the problem by the example of using information from a map and measurements from a rangefinder that measures distances to walls in a specified direction. The problem is formulated in the context of nonlinear filtering theory, by analogy with a similar problem of correcting the navigation system data using information on the coastline [8]. Under this statement, we propose an algorithm for the problem solution based on the approximation of posterior p.d.f. with the use of the point-mass method, which allows us to calculate both the estimates and their respective accuracy characteristics. The simulation results illustrating the advantages of the problem solution on the proposed basis are given.

## Problem statement

Consider a mobile object within an indoor environment, with the object's position and heading denoted as $X^o = (X_1^o, X_2^o)^T$ and $K$ , correspondingly. It is assumed, for simplicity, that

- the object has a zero velocity,
- the object is located on a horizontal plane in a rectangular room with width $L_1$ and length $L_2$ as shown in Fig.1,
- the position of the range finder coincides with the object's position.

For definiteness we assume that the object has a laser rangefinder (LRF) that provides ranges (distances) to the nearest obstacle in direction $K_i = K + (i-1)\Delta k$ , where $\Delta k = 0.36^\circ$ is the LRF resolution [9].

Introduce function $\rho_i(X^o, K_i)$ to determine the range between the object and the room wall in the specified direction $K_i$ . In general, this function has the following form:

$$\rho_i(X^o, K_i) = \sqrt{\left(X_1^{Ci} - X_1^o\right)^2 + \left(X_2^{Ci} - X_2^o\right)^2}\,, \qquad (1)$$

---

1 Professor, Doctor of Sciences
2 PhD, Researcher

where $X_1^{Ci}, X_2^{Ci}$ are the coordinates of the cross points of the LRF beam in the direction of $K_i = K + (i-1)\Delta k$ and the straight line representing the corresponding wall. The equation for the straight lines describing the walls is easy to obtain as an equation of a line passing through two given points. To determine function (1) for fixed values of $X^o$ and $K_i$, it is necessary, first, to choose, for every $K_i$, the appropriate walls to which the ranges should be found, then to find the coordinates of these cross points, and finally select the special point $X_1^{Ci}$, $X_2^{Ci}$ with the least distance to be used in Equation (1).

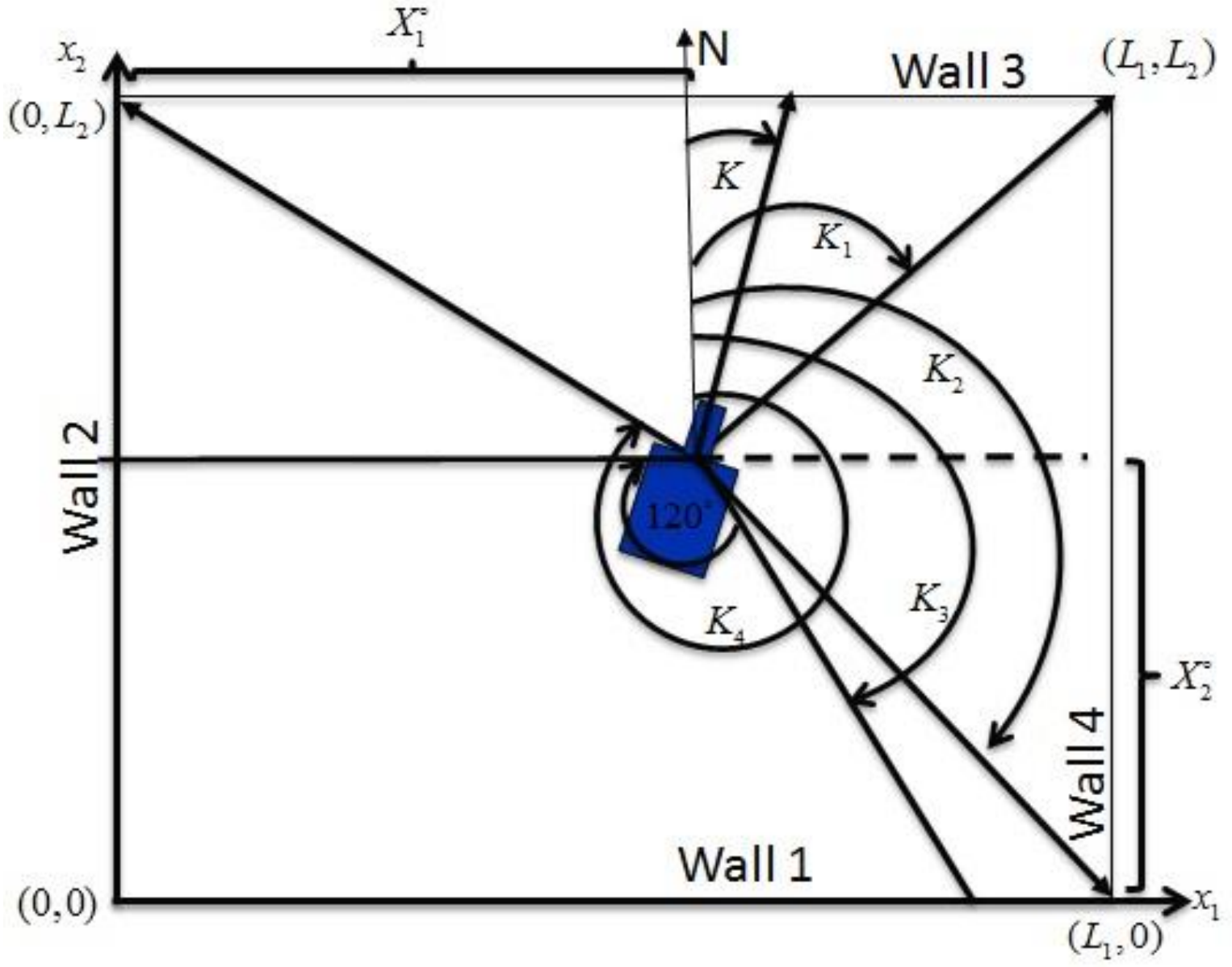

Fig. 1 Example of an object's location on the horizontal plane in a rectangular room with width $L_1$ and length $L_2$.

The example of different formulas that should be used in equation (1) according to the above figure is given below.

$$\rho(X^\circ, K_i) = \begin{cases} \dfrac{L_2 - X_2^\circ}{\cos(K_i)}, K_i \in (K_4, K_1)\ \forall K_i = K \pm i\Delta K\ , \\ \dfrac{L_1 - X_1^\circ}{\sin(K_i)}, K_i \in (K_1, K_2)\ \forall K_i = K + i\Delta K\ , \\ \dfrac{X_2^\circ}{|\cos(K_i)|}, K_i \in (K_2, K_3)\ \forall K_i = K + i\Delta K, \\ 0, K_i \in (K_3, 120 + K_3)\ \forall K_i = K + i\Delta K, \\ \dfrac{X_1^\circ}{|\sin(K_i)|}, K_i \in (120 + K_3, K_4)\ \forall K_i = K + i\Delta K, \end{cases} \qquad \left. \begin{aligned} & K_1 = \arctan(\frac{L_1 - X_1^\circ}{L_2 - X_2^\circ}), \\ & K_2 = \pi\text{-}\arctan(\frac{L_1 - X_1^\circ}{X_2^\circ}), \\ & K_3 = 120 + K, \\ & K_4 = 2\pi - \arctan(\frac{X_1^\circ}{L_2 - X_2^\circ}). \end{aligned} \right\}$$

(2)

Introduce state vector $x = \left[X_1^o,\quad X_2^o,\quad K\right]^T = \left[x_1,\quad x_2,\quad x_3\right]^T$ and formulate the following problem: to estimate state vector

$$\dot{x} = 0 \qquad (3)$$

using range measurements given as follows:

$$y_i = \rho_i(x) + v_i\,, \qquad (4)$$

where $\rho_i(x) = \rho_i(X^o, K_i)$ is a nonlinear function (1); $v_i$ are range measurement errors with known statistical characteristics. It is assumed that vector $x$ is a random vector with a prior probability density function (p.d.f.) $f(x)$. The estimation problem can be formulated using the Bayesian approach. For navigation problems, the traditional RMS (root-mean-square) criterion in the form of the unconditional covariance matrix of the estimation error of vector $x$ is chosen. This matrix is determined as follows:

$$G_i = E_{x,Y_i}\{(x - \hat{x}(Y_i))(x - \hat{x}(Y_i))^T\}\,. \qquad (5)$$

In this case, optimal estimate $\hat{x}(Y_i)$ can be obtained as the value that minimizes the unconditional covariance matrix (5). It is worth recalling that matrix minimization is usually understood as minimization of its corresponding quadratic form. As is known, this estimate is determined as a conditional mathematical expectation, i.e., [10]:

$$\hat{x}_i(Y_i) = \int x f(x/Y_i)dx, \tag{6}$$

and its corresponding conditional covariance matrix is determined as follows:

$$P_i^x = \int (x - \hat{x}_i(Y_i))(x - \hat{x}_i(Y_i))^T f(x/Y_i)dx, \tag{7}$$

where $Y_i = (y_1, y_2, ...y_i)^T$ is the vector including a set of measurements accumulated by the current moment of time, $f(x/Y_i)$ is the posterior (conditional) p.d.f. given by the following equation:

$$f(x/Y_i) = \frac{f(Y_i/x)f(x)}{\int f(Y_i/x)f(x)dx}, \tag{8}$$

where $f(Y_i/x)$ is the likelihood function. For simplicity, range measurement errors are assumed to be independent on each other and they are zero-mean random variables with the Gaussian p.d.f. $N(v_i;0,r_i^2)$ with $E(v_i v_j) = 0, \forall i \neq j$. The likelihood function can be represented as follows:

$$f(Y_i/x) = c \exp\left\{-\frac{1}{2}\sum_{j=1}^{i}\frac{\left(y_j - \rho_j(x)\right)^2}{r_j^2}\right\}, \tag{9}$$

where $c$ is a normalization factor .

It should be noted that in (6), (7) the integrals are considered as multidimensional integrals with infinite limits and in (9) function $\rho_j(x)$ is defined according to (1), where its arguments are the values of $x$ belonging to the uncertainty domain. It is easy to notice that the proposed problem statement completely coincides with the problem statement of map-aided navigation solved in the context of Bayesian estimation using nonlinear filtering theory. Thus it makes sense to use the experience gained in map-aided navigation methods to design estimation algorithms for the problem under study [10-13].

**Optimal algorithm for estimation**

A wide variety of algorithms for nonlinear filtering estimation problems have been designed to meet the needs of posterior p.d.f. approximation [10]. It should be noted that for the analysis of the capability of a particular navigation method for the problem under study the type of the algorithm used is of little significance, but it is quite important to be sure that the chosen method for approximation will calculate the integrals in (6), (7) with the required accuracy. In this regard, both the point-mass method and the Monte Carlo method can be used to develop an algorithm capable of providing an optimal estimate and its corresponding covariance matrix. Note that recently preference has been given to Monte Carlo method or its modifications, such as particle filter [2, 5]. However, it is worth emphasizing that for the given problem with invariable state vector (zero-velocity object), both methods are similar in their capabilities and properties [14]. In this paper, we use the point-mass method that was first proposed in [15].

To illustrate how this method works, assume, for simplicity, that the heading of the object is known and it is required to estimate the coordinates only. It is supposed that the prior uncertainty domain has a center at the room center and its dimensions coincide with the room itself. For prior p.d.f. the following approximation is used:

$$f(x_1, x_2) = f(x_1)f(x_2) = \frac{1}{MN}\sum_{l=1}^{N}\delta(x_2 - x_2^n)\sum_{j=1}^{M}\delta(x_1 - x_1^l), \tag{10}$$

where $x_1^l$, $x_2^j$ are the grid points determining possible values of the coordinates , $n = \overline{1.N}$, $j = \overline{1.M}$.

Substituting (10) in (8), the following approximation for p.d.f can be obtained:

$$f(x_1, x_2/K, Y_i) = \sum_{n=1}^{N}\sum_{l=1}^{M}\mu_i^{nl}\delta(x_1 - x_1^n)\delta(x_2 - x_2^l), \qquad \mu_i^{nl} = \frac{\tilde{\mu}_i^{nl}}{\sum_{n=1}^{N}\sum_{l=1}^{M}\tilde{\mu}_i^{nl}}, \tag{11}$$

where

$$\tilde{\mu}_i^{nl} = \exp\left\{-\frac{1}{2}\sum_{j=1}^{i}\frac{\left(y_j - \rho_j(x^n, x^l, K)\right)^2}{r_j^2}\right\}. \tag{12}$$

Using (11), (12), it is easy to obtain the estimates and the covariance matrix (7). In particular, the estimates and the diagonal elements of matrix (7) can be obtained as follows:

$$\hat{x}_{i1} = \sum_{n=1}^{N}\sum_{l=1}^{M}x_1^n\mu_i^{nl}, \hat{x}_{i2} = \sum_{n=1}^{N}\sum_{l=1}^{M}x_2^l\mu_i^{nl}, \tag{13}$$

$$P_i[1,1] = \sum_{n=1}^{N}\sum_{l=1}^{M}\left(x_1^n\right)^2\mu_i^{nl} - \left(\hat{x}_{i1}\right)^2, P_i[2,2] = \sum_{n=1}^{N}\sum_{l=1}^{M}\left(x_1^l\right)^2\mu_i^{nl} - \left(\hat{x}_{i2}\right)^2. \tag{14}$$

As stated in the introduction, to know how to obtain accuracy characteristics in the form of a covariance matrix is very important in solving the problems of navigation data processing. Here, the necessary characteristics are easy to obtain using (14).

The presence of an algorithm for calculating an optimal estimate makes it possible to calculate unconditional covariance matrix (5) which characterizes the potential estimation accuracy of an estimate that can be also used for evaluating the effectiveness of different simplified algorithms. Matrix (5) is usually calculated with the use of the Monte Carlo method, which is implemented by multiple solutions of the estimation problem [16].

Consider an example illustrating the described procedures and the possibility of using them to solve the navigation problem using information from a map and a range finder.

**Example**

Assume that an object equipped with an LRF is located at the center of a room with dimensions L1=4m and L2=6m, its heading being K=20°. The prior p.f.d. is given as a uniform distribution with the dimensions that coincide with the room dimensions, so that $\sigma_{01} = 1.1$ m and $\sigma_{02} = 1.7$ m . The RMS values of measurement errors are assumed similar and equal to 0.05m. Using the approximation in (10), we can obtain optimal estimates and their corresponding RMS values $\sqrt{P_i[1,1]}$ , $\sqrt{P_i[2,2]}$ . Three measurements were taken at $K_1$=326.3° (the first measurement corresponds to the upper left angle of the room), at $K_2 = 0$ (the second measurement corresponds to the measurement perpendicular to wall 3 ) and at $K_3$=33.7° (the third measurement corresponds to the upper right angle of the room). The results obtained for different combinations of these three measurements and their corresponding graphs of the posterior p.d.f. are given in Figs. 2, 3, and in Table 1.

This is how we can comment briefly on the results. The posterior p.d.f. has a complicated form but an explicable one. By processing just one measurement, the posterior p.d.f. is determined by the isoline of the positions corresponding to the fixed range value in the given direction determined by the range function in (1). As for the second measurement, these isolines are parallel to wall 3, whereas in the case of the first and third measurements, they take the forms as described in Fig. 2.a1 and Fig. 2.a3, respectively. Also, it is clear why using the first, second and third measurements together, as in Fig. 3, can significantly reduce the prior uncertainty domain for two coordinates at a time. This occurs because every likelihood function, corresponding to these measurements, has a nonzero value at different regions (values) of $x_1, x_2$ and only around the real position of the object, these regions adjoin each other.

The results given in Fig.2 and Table 1 show that, in general, the current accuracy characteristics $\sqrt{P_i[1,1]}$ $\sqrt{P_i[1,1]}$ correctly reflect the accuracy of the obtained estimates. The analysis of the behavior of the posterior p.d.f. and its corresponding accuracy creates preconditions for planning a choice of measurements that are of primary importance in determining position. Also, it is clear that by using the described algorithm and the Monte Carlo method, we can implement the procedure of calculating the unconditional covariance matrix to estimate the effectiveness of this type of simplified algorithms that are used to solve the problem under study.

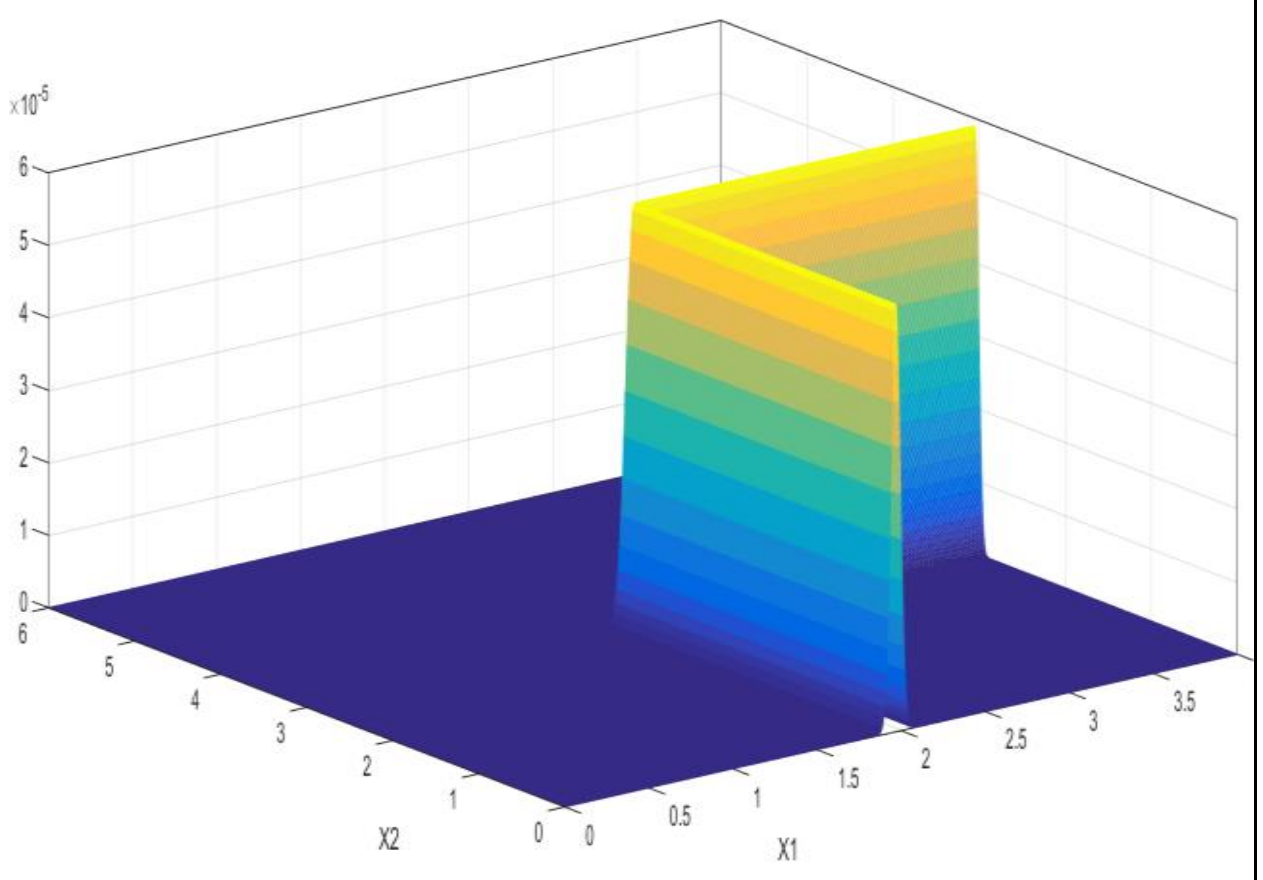

a.1. Posterior p.d.f. for the first measurement at $K_1$=326.3°, the upper left angle of the room.

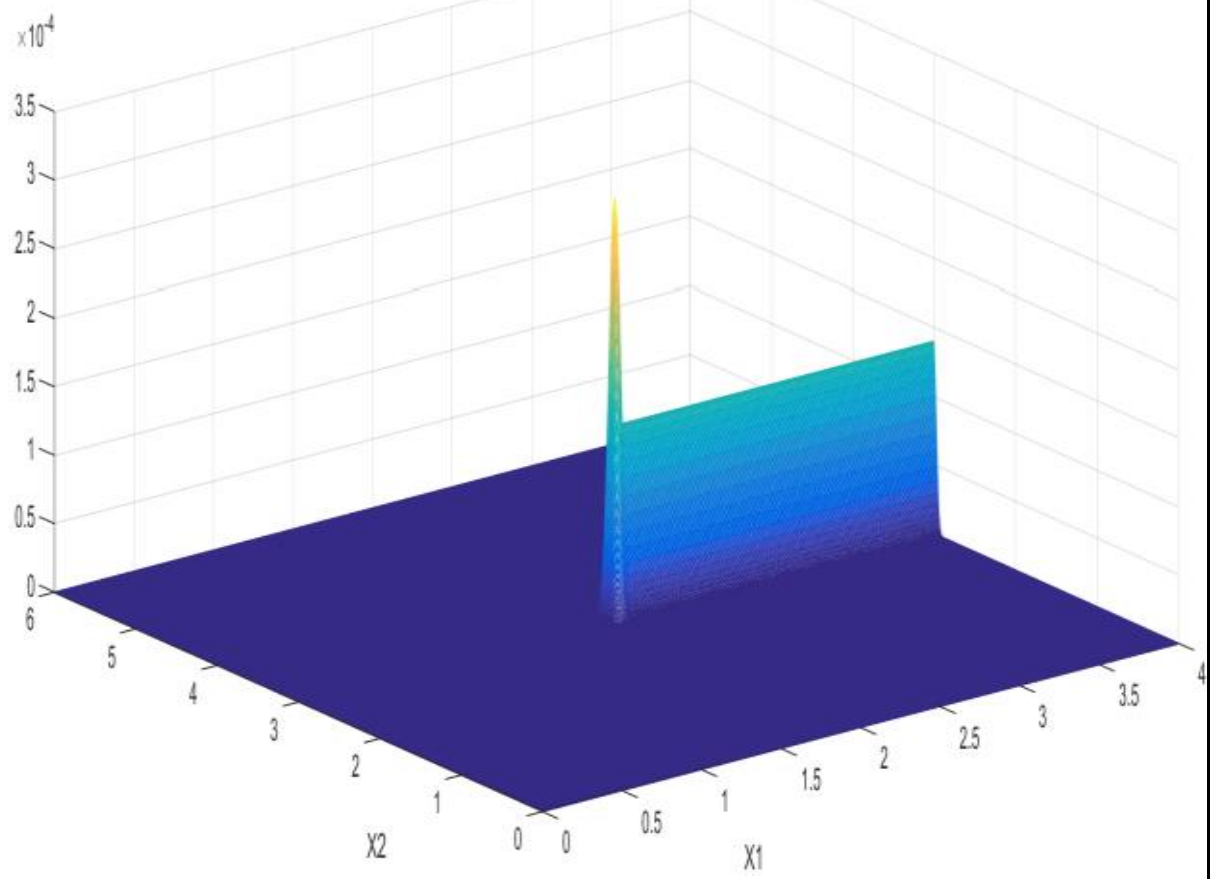

b.1. Posterior p.d.f. for the first and second measurement at $K_1$=326.3° and $K_2$=0°, respectively.

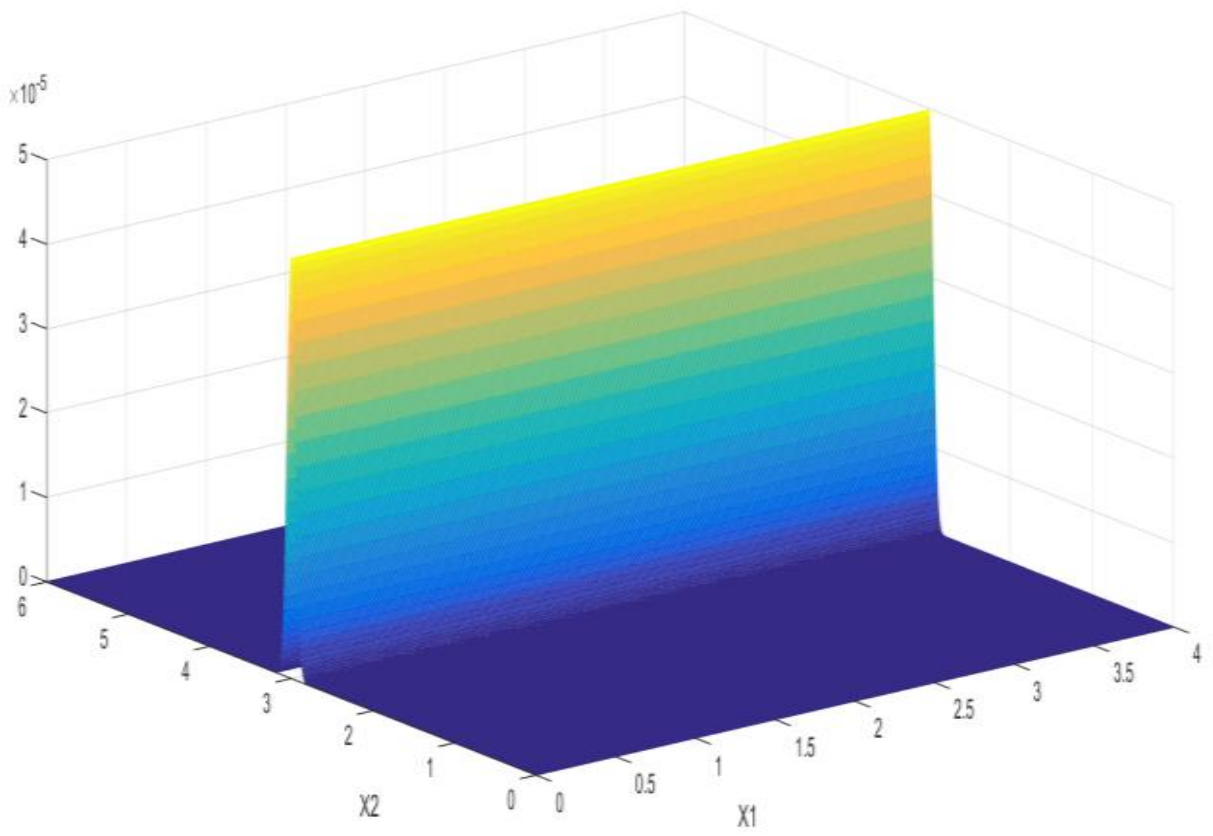

a.2. Posterior p.d.f. for the second measurement at $K_2$=0.

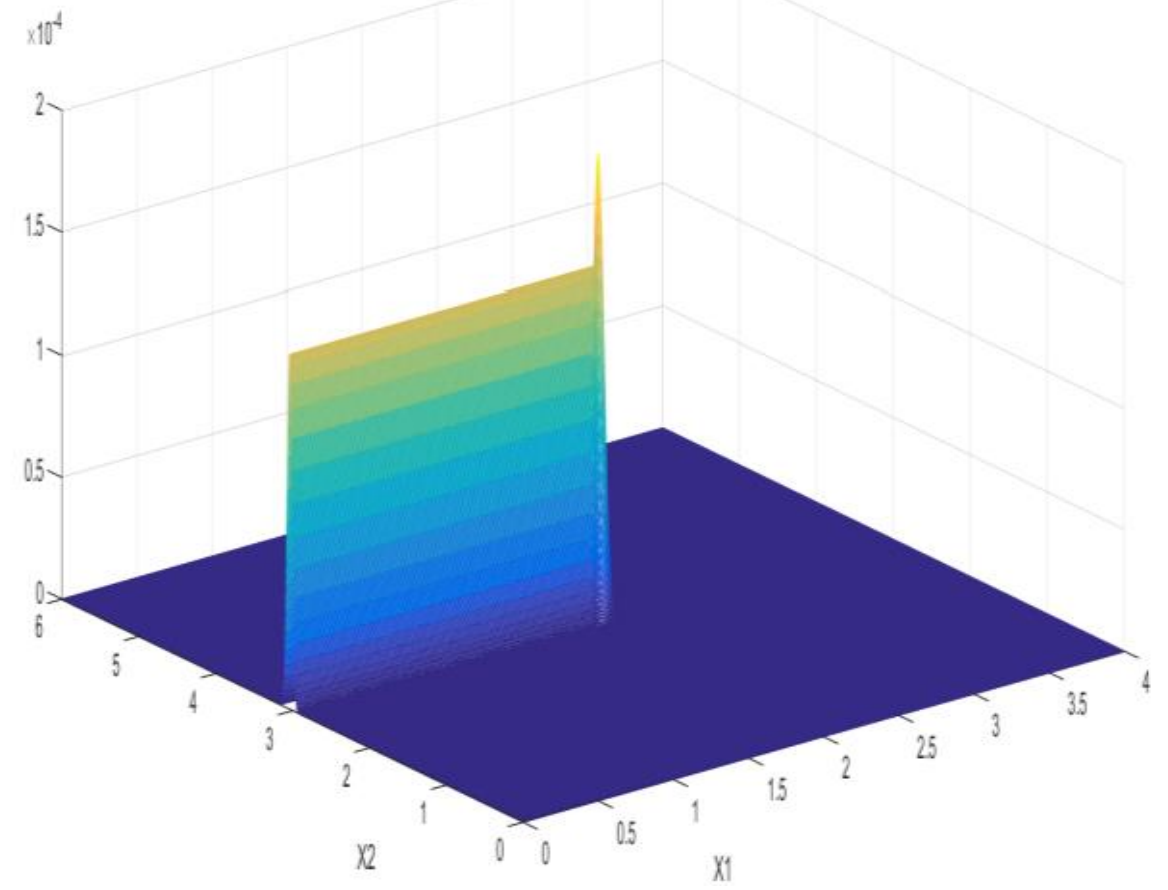

b.2. Posterior p.d.f. for the second and third measurements at $K_2$=0°and $K_3$=33.7°, respectively.

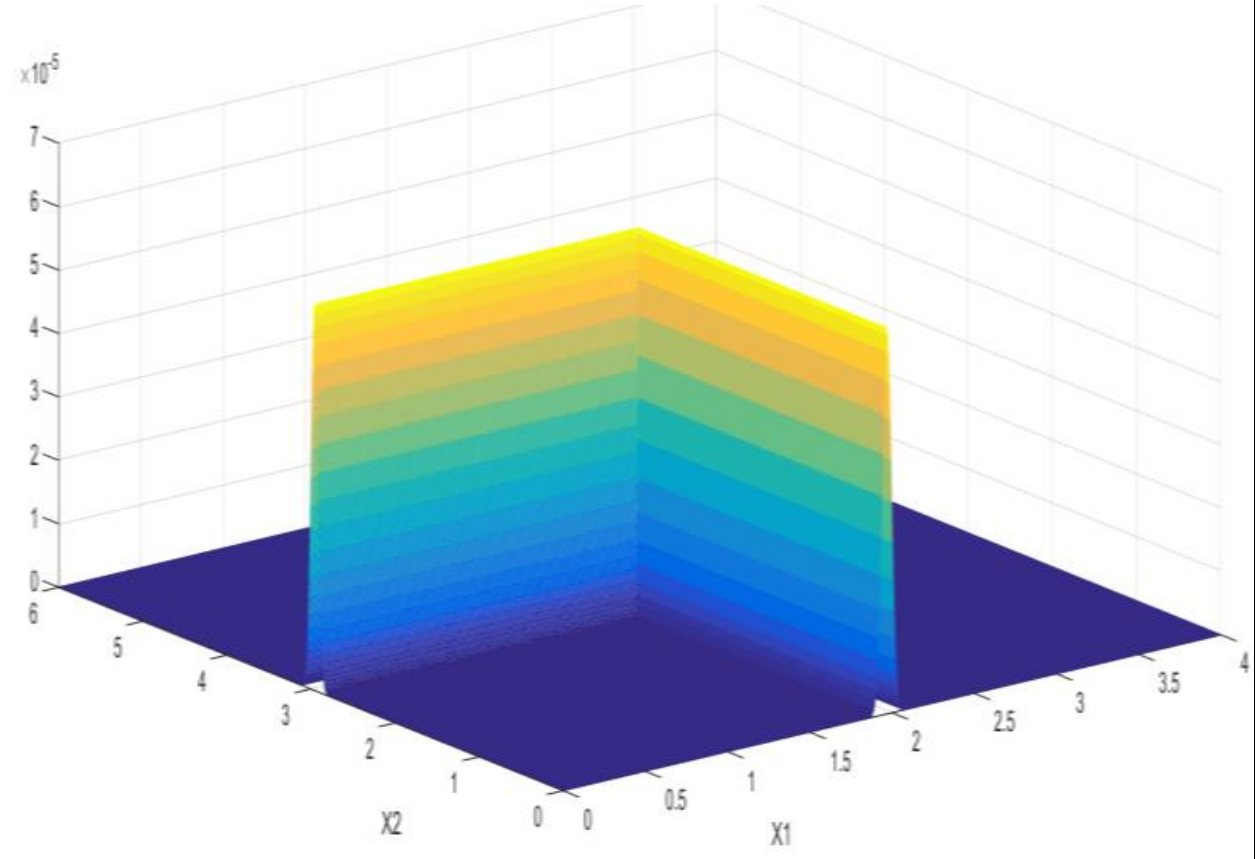

a.3. Posterior p.d.f. for the third measurement at $K_3$=33.7°, the upper right angle of the room.

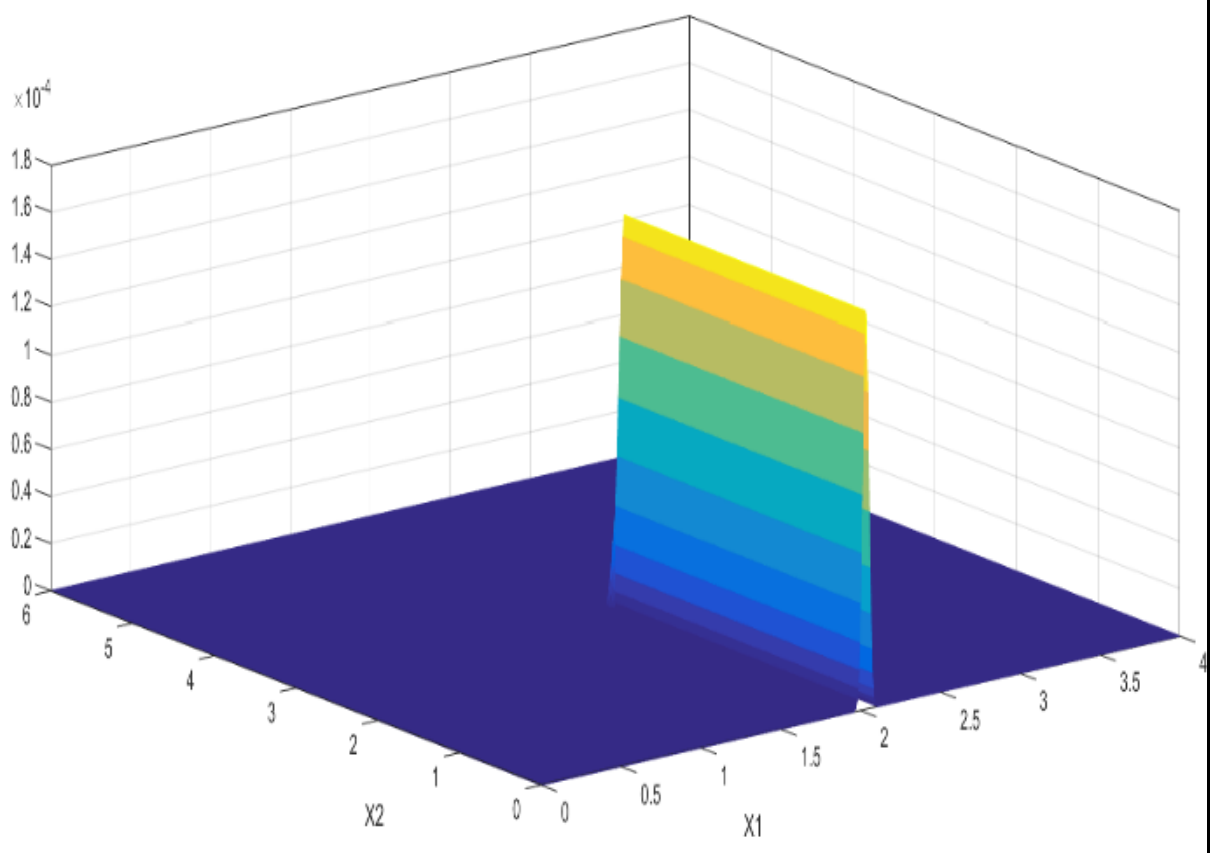

b.3. Posterior p.d.f. for the first and third measurements at $K_1$=326.3°and $K_3$=33.7°, respectively.

Fig. 2. Posterior p.d.f. for different combinations of measurements

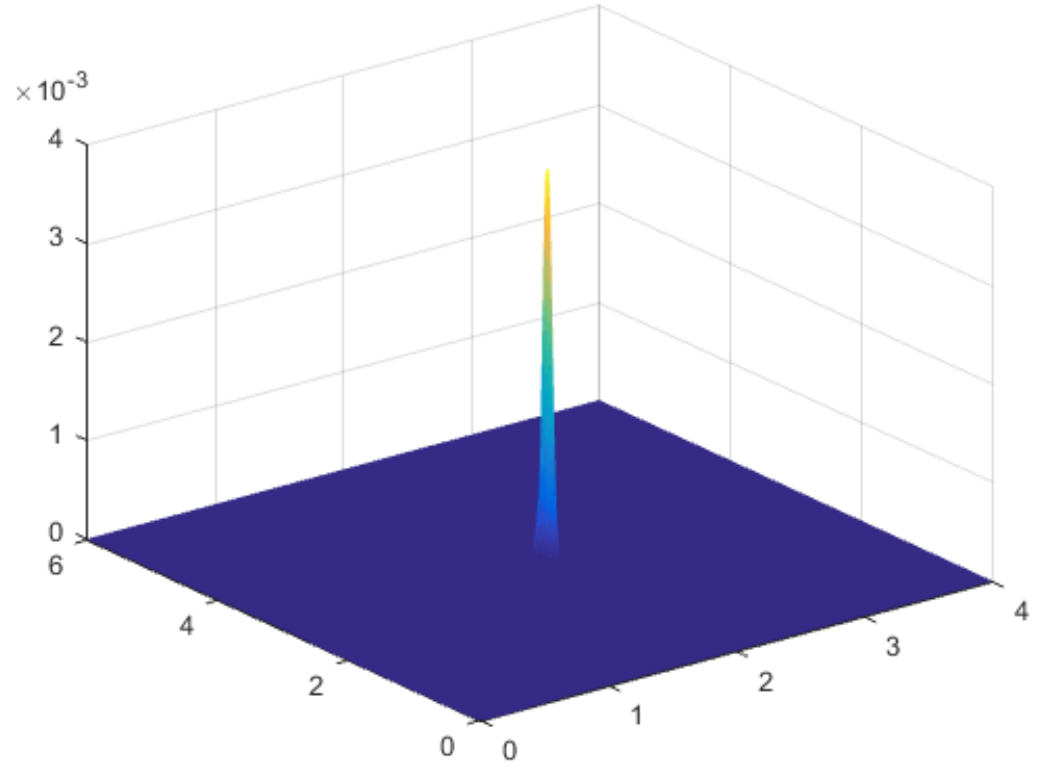

Fig. 3. Posterior p.d.f. after using the three measurements.

| **Measurements** | **1** | **2** | **3** | **1+2** | **2+3** | **1+3** | **1+2+3** |
|---|---|---|---|---|---|---|---|
| $\sqrt{P_i[1,1]}$ | 0.64 | 1.1 | 0.64 | 0.63 | 0.61 | 0,02 | 0,02 |
| $\sqrt{P_i[2,2]}$ | 0.96 | 0.05 | 0.96 | 0.03 | 0.03 | 0.87 | 0.03 |

Table 1. RMS value of the position estimate errors using different combinations of measurements.

Speaking about the future work, we think that it is important to investigate the proposed algorithm for an object with nonzero velocity and an unknown heading value. Besides, we should refine the error model for the sensors used [17]. It will be also very useful to investigate the possibility of using the Rao-Cramer inequality to analyze the accuracy of the solution of the problem under study as described in [18, 19].

## Conclusions

The problem of determining position and heading of a mobile object within an indoor environment using a map and measurements of distances to the walls is formulated in the context of nonlinear filtering theory.

The proposed problem statement offers the following main advantages: on the one hand, it creates preconditions for analysis of the potential of the navigation method used, and, on the other hand, it allows obtaining not only the sought estimates, but the current accuracy characteristics as well, which is of critical importance in navigation problems.

The algorithm for the problem solution is presented, based on approximation of posterior p.d.f. with the use of the point-mass method. A numerical example is given to illustrate the advantages of the proposed problem statement and the resultant algorithm for its solution.

This work was supported by the Russian Foundation for Basic Research, project №14-08-00347.